\pdfoutput=1

\documentclass[11pt]{article}

\usepackage[final]{acl}
\usepackage{tcolorbox}
\usepackage{times}
\usepackage{latexsym}
\usepackage[T1]{fontenc}
\usepackage[utf8]{inputenc}
\usepackage{microtype}
\usepackage{float}
\usepackage{xcolor}
\usepackage{inconsolata}
\usepackage{hyperref}
\usepackage{graphicx}
\usepackage{booktabs}
\usepackage{multirow} 
\usepackage{comment}
\usepackage[english]{babel}
\usepackage{csquotes}
\usepackage{colortbl}
\usepackage{pifont}
\DeclareUnicodeCharacter{0266B}{\ding{74}}

\definecolor{lightgray}{gray}{0.9}

\title{\ours{}: Syntax Tree-Guided Retrieval and Reasoning \\ for Knowledge-Intensive Question Answering}

\author{Boyi Zhang \\
  University of Rochester \\
  \texttt{bzhang44@u.rochester.edu } \\\And
  Zhuo Liu \\
  University of Rochester \\
  \texttt{zliu106@ur.rochester.edu} \\\And
  Hangfeng He \\
  University of Rochester \\
  \texttt{hangfeng.he@rochester.edu} \\}

\begin{document}
\maketitle
\begin{abstract}
In real practice, questions are typically complex and knowledge-intensive, requiring Large Language Models (LLMs) to recognize the multifaceted nature of the question and reason across multiple information sources. Iterative and adaptive retrieval, where LLMs decide \textit{when and what to retrieve} based on their reasoning, has been shown to be a promising approach to resolve complex, knowledge-intensive questions. However, the performance of such retrieval frameworks is limited by the accumulation of reasoning errors and misaligned retrieval results. To overcome these limitations, we propose \ours{} (Syntax \textbf{Tree}-Guided \textbf{R}etrieval \textbf{a}nd \textbf{Re}asoning), a framework that utilizes syntax trees to guide information retrieval and reasoning for question answering. Following the principle of compositionality, \ours{} traverses the syntax tree in a bottom-up fashion, and in each node, it generates subcomponent-based queries and retrieves relevant passages to resolve localized uncertainty. A subcomponent question answering module then synthesizes these passages into concise, context-aware evidence. Finally, \ours{} aggregates the evidence across the tree to form a final answer. Experiments across five question answering datasets involving ambiguous or multi-hop reasoning demonstrate that \ours{} achieves substantial improvements over existing state-of-the-art methods.\footnote{Our code is publicly available at \href{https://github.com/billycrapediem/TreeRare}{https://github.com/billycrapediem/TreeRare}.}

\end{abstract}

\section{Introduction}
\begin{figure}[t]
    \centering
    \includegraphics[width=1\linewidth]{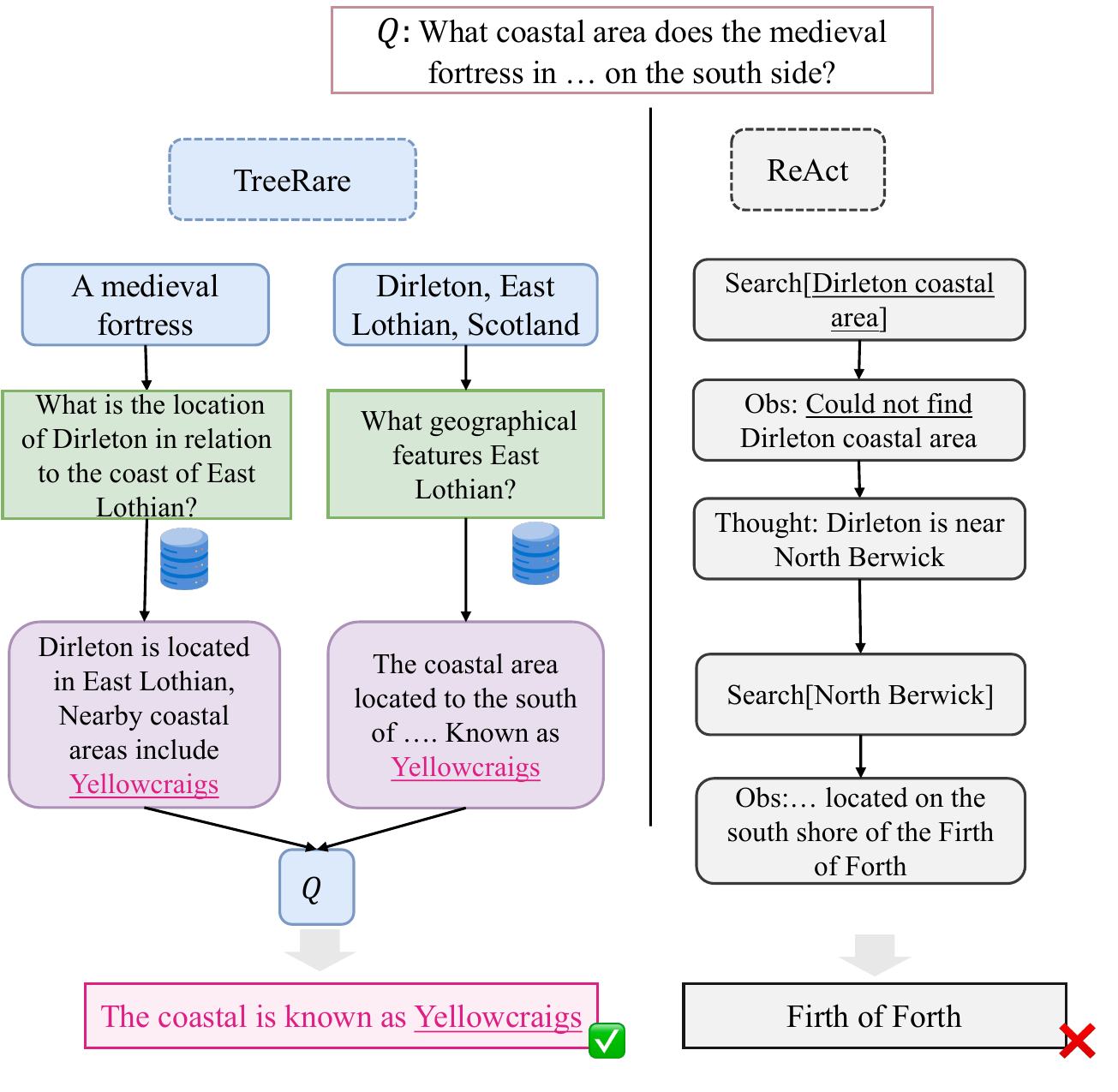}
    \caption{Comparison of \ours{} and ReAct\cite{yao2022react} on a multihop question. \ours{} decomposes the question into structured sub-questions and retrieves focused evidence. In contrast, ReAct fails to generate useful query and misidentifies the region. }
    \label{fig:teaser}
\end{figure}
Large Language Models (LLMs) \cite{chowdhery2023palm,achiam2023gpt} have demonstrated remarkable capabilities across a wide range of natural language processing (NLP) tasks, including text generation and question answering, often achieving strong performance in few-shot or even zero-shot settings without task-specific fine-tuning \cite{NEURIPS2020_1457c0d6}. Despite their impressive performance, LLMs generate plausible but factually incorrect statements due to over-reliance on their parametric knowledge when tackling knowledge-intensive tasks that demand factual accuracy and external grounding \cite{roberts-etal-2020-much,maynez-etal-2020-faithfulness, chen2022towards}. To address this issue, existing work has shown that retrieval-augmented generation (RAG) can largely reduce factual hallucination by incorporating LLMs with external knowledge sources \cite{lewis2020retrieval,guu2020retrieval}.

RAG enhances LLMs by integrating external knowledge retrieval into the generation process. Traditional RAG systems follow a ``retrieve-then-read'' paradigm, where a retriever selects top-k documents based on similarity metrics, and the LLM generates responses conditioned on retrieved documents \cite{lewis2020retrieval}. While effective for well-posed queries, this paradigm often struggles to answer ambiguous or multihop questions, as similarity-based retrieval may miss relevant evidence. To address this limitation, many works enhance retrieval quality via retrieval judgment, adaptive search mechanisms, or question decomposition to improve retrieval quality and performance of complex knowledge-intensive tasks. \cite{li-etal-2025-topology,asai2024selfrag,tan-etal-2024-small,yao2022react} However, as illustrated in Figure~\ref{fig:teaser}, these methods often rely on internal reasoning in LLMs, which can cause errors to accumulate across steps. Mistakes made during reasoning or retrieval can result in information that is misaligned with the original intent of the question, leading to noisy inputs and incorrect final answers \cite{li2024chainofknowledge}.

Inspired by the principle of compositionality —\textit{the meanings of complex expressions are constructed from the meanings of the less complex expressions that are their constituents} \cite{fodor2002compositionality}— we then ask: Can the syntactic structure of complex, knowledge-intensive questions guide effective retrieval and inference toward correct answers? We leverage syntax trees as a basis for question decomposition, since parsing tree has been shown to be effective in capturing the syntactic relations between each phrase in a sentence \cite{li-etal-2015-tree}. Then, we propose a bottom-up traversal of the syntax tree, where each child node is processed first, and its output is used to guide the processing of its parent node. Additionally, observing that LLMs frequently fail to detect ambiguity or knowledge gaps present within sub-phrases \cite{piryani-etal-2024-detecting,kim-etal-2024-aligning}, we provide LLMs at each node with information from the child nodes to formulate queries that resolve the associated sub-phrase uncertainty. Furthermore, recognizing that LLM performance degrades when conditioned on long or noisy inputs \cite{liu-etal-2024-lost,xu2024retrieval}, we introduce a subcomponent question answering that synthesize the retrieved context into concise, phrase-relevant evidence. 

\begin{figure*}[t]
    \centering
    \includegraphics[width=1\linewidth]{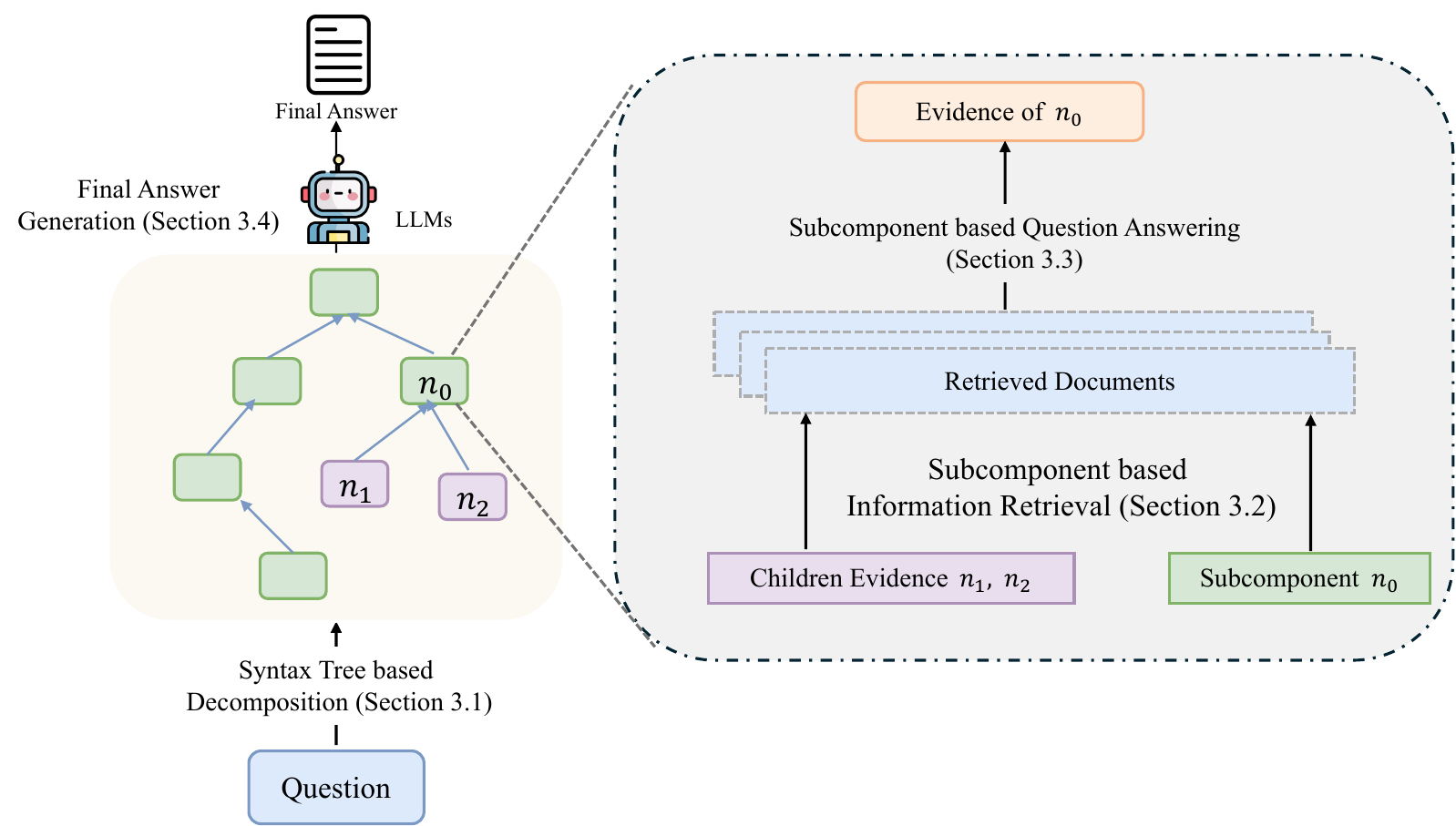}
    \caption{\textbf{Overview of the \ours{} framework}. Given a knowledge-intensive question, \ours{} first decomposes it into a syntax tree. It then traverses the tree in a bottom-up manner. At each node, the framework generates subcomponent-based queries conditioned on evidence from its child nodes and the current phrase. These queries guide document retrieval, and a subcomponent question answering module extracts evidence related to the generated queries. Finally, all node-level evidence is aggregated to produce the final answer. }
    \label{fig:main}
\end{figure*}

Combining these modules, we propose syntax tree-guided retrieval and reasoning (\textbf{\ours{}}). \ours{} incrementally retrieves and resolves subcomponents of a question in accordance with its syntax structure. As illustrated in Figure~\ref{fig:main}, \ours{} traverses the syntax tree in a bottom-up manner, resolving the uncertainty at each node through a two-stage process, starting with subcomponent-based retrieval, followed by subcomponent question answering. Upon completing the traversal, \ours{} constructs targeted, comprehensive evidence for each sub-phrase in the parsing tree and aggregates these evidence across nodes to generate a final answer.

Our contributions are as follows: (1) We propose \ours{} to handle complex, knowledge-intensive questions, which enhances LLMs' performance by interleaving retrieval with reasoning over the syntax tree. 
(2) We introduce retrieval-only counterpart, Tree-Retrieval, effectively improving the retrieval quality without involving any LLM reasoning.  (3) We perform experiments across multiple multihop and ambiguous question answering (QA) benchmarks for three LLM backbones. On multi-hop QA benchmarks, \ours{} achieves an average relative improvement up to $17.8\%$. For ambiguous QA, \ours{} yields an average improvement of $23.7\%$ across various evaluation metrics.

\section{Related Work}
\subsection{LLMs for Reasoning}
Significant efforts have been dedicated to enhance the reasoning capabilities of LLMs during the inference phase. Chain-of-prompting (CoT) \cite{wei2022chain} prompting introduces intermediate reasoning steps between the initial query and the final answer, thereby improving performance on complex tasks. Building upon this, self-consistency (SC) \cite{wang2022self} generates multiple reasoning paths and employ majority voting to select the final answer; self-verification \cite{shinn2023reflexion} prompts LLMs to reflect on their outputs and iteratively refine them through feedback. Additionally, Tree-of-Thought (ToT) and Reasoning via Planning (RAP) prompting \cite{yao2023tree, hao2023reasoning} extend the CoT approach. These methods further enhance LLMs’ reasoning abilities by exploring multiple reasoning paths with different tree search algorithm. Above approaches rely solely on the internal reasoning traces of LLMs and lack structural guidance, such as syntax tree included in \ours{}.

\subsection{Retrieval Augmented Generation}
Common RAG systems proceed in a retriever-then-read paradigm, where it first retrieves relevant documents based on the user's query using either sparse or dense retrieval and then takes the retrieved information in addition to the question as input to generate final answer \cite{Khandelwal2020Generalization,shi-etal-2024-replug,borgeaud2022improving}. This paradigm outperforms regular LLMs, especially for the knowledge-intensive single-hop questions. In order to answer these single-hop questions, the needed information is evident from the question itself, such that a one-time retrieval can find the documents that contain the answer \cite{trivedi-etal-2022-musique}. 

However, this paradigm is inadequate for complex, knowledge-intensive questions, such as multihop or ambiguous questions. Standard RAG systems retrieve documents based solely on the original query, without accounting for the evolving information needs of further reasoning steps \cite{talmor-berant-2018-web,amouyal-etal-2023-qampari}. In the case of ambiguous questions, such approaches run the risk of making an early commitment to a single interpretation and potentially overlooking alternative meanings necessary for accurate comprehension. \cite{lee2024ambigdocs,gao-etal-2021-answering}. 
Notably, LLMs have demonstrated their strong ability in decomposing complex tasks into different sub queries to facilitate its performance on complex, knowledge-intensive tasks \cite{drozdov2022compositional,khot2023decomposed,dua-etal-2022-successive}. Several approaches incorporate LLMs' generating contents, including intermediate reasoning steps or heuristic answers, to guide retrieval \cite{he2022rethinking,trivedi-etal-2023-interleaving,tan-etal-2024-small}. Similar to how humans iteratively resolve complex questions by identifying salient information gaps, querying on search engine, and progressively narrowing uncertainty until reaching a final answer, further work has applied LLM agents for information retrieval by leveraging LLMs’ reasoning capabilities to dynamically determine both \textit{when and what to retrieve} \cite{jiang-etal-2023-active,li-etal-2025-topology,yao2022react}. While these adaptive RAGs depend on reasoning traces or heuristic decision-making, \ours{} utilizes the syntactic structure of the question to guide the retrieval and reasoning.
\section{\ours{}: Syntax Tree-Guided Retrieval and Reasoning}

In this section, we give a detailed explanation of \ours{}. Our approach is built on three key intuitions: (1) answering complex reasoning questions requires addressing uncertainty within each phrase of a question; (2) the uncertainty associated with a phrase depends on clarifying its constituent sub-phrases, as understanding the parts is necessary to resolve the whole; and (3) effective retrieval should target diverse and fine-grained evidence that aligns with the phrase. 

\subsection{Syntax Tree-Based Decomposition}
Given an input question $Q$, \ours{} converts it into its corresponding syntax tree. The resulting tree comprises a set of nodes $N$, where each node $n \in N$ corresponds to a sub-phrase $s_{n}$ and is associated with a set of child nodes $C_{n}$. Specifically, $s_{n}$ spans all tokens dominated by node $n$ in the parse tree. These syntax structures make explicit the constituency-based or dependency-based relationships between phrases, displaying how meaning is composed from sub-units of the question. \ours{} interprets each sub-phrase as a constraint of the question and reasons over them incrementally. Following principles of compositional semantics \cite{fodor2002compositionality}, the reasoning process proceeds in a bottom-up tree traversal, ensuring that node $n$ is processed only after resolving all its children $C_{n}$. Such ordering provides a basis for constructing a stepwise reasoning path grounded in syntax decomposition. 
\subsection{Subcomponent-Based Information Retrieval}\label{method:question_generation}
At each node $n$ in the syntax tree, \ours{} generates a set of subcomponent-based queries to address the latent knowledge gaps associated with its sub-phrase $s_{n}$. Complicated uncertainty often emerges not at the level of individual sub-phrases, but from their interaction. For instance, through the composition of sub-phrases, novel entities may emerge, and modifiers can introduce ambiguity or context-sensitive reinterpretations. Therefore, even if each child node contributes reliable evidence, $E_c$, this evidence alone may be insufficient to resolve the uncertainty within $s_n$. To bridge these gaps, \ours{} prompts the LLM to generate multiple subcomponent-based queries that are conditioned on global question, local sub-phrase and evidence from its children. Formally, the query set is constructed as: $Q_{n}=QG(Q,s_{n},\left\{ E_c|c \in C_{n} \right\})$, where $QG$ is a function that prompts LLMs to generate a set of simple queries. This formulation ensures two key properties of generated queries:
\begin{itemize}
    \item \textbf{Compositional Grounding}: Queries are crafted to resolve the information gap that emerges from interactions between sub-phrases or from novel information introduced through their composition.
    \item \textbf{Explicit Reasoning}: Queries function as an intermediate reasoning step to resolve $s_n$, which eventually leads to coherent inference toward the full question.
\end{itemize}
Following the generation of subcomponent queries, \ours{} initiates a targeted retrieval procedure aimed at acquiring external textual documents that directly resolve the subcomponent-based queries, $Q_n$. Each query $q\in Q_n$ guides the retriever over a large corpus to obtain a set of top-ranked documents $d_{n,q}$. The complete retrieved context for node $n$ is then defined as $D_{n}=\bigcup_{q \in Q_n} d_{n,q}$. 

\subsection{Subcomponent Question Answering }\label{method:information_filter}
 Naively combining all retrieved documents leads to excessive input length and noise, especially harmful under the "Lost-in-the-Middle" effect in LLMs \cite{liu-etal-2024-lost}. To mitigate this effect, \ours{} introduces subcomponent question answering module. It processes the retrieved content to retain information that is salient to sub-phrase $s_n$. At the same time, it tries to address remaining reasoning gaps that arise when integrating evidence from the subcomponents. These may include contradictions, underspecified relationships, or missing inferences needed to represent the full meaning of $s_n$ from its children. Formally, given generated queries $Q_{n}$ and retrieved documents $D_n$, the LLM is instructed to produce a concise set of answers. $E_{n}=SAG(Q_{n},D_{n})$, where $SAG$ is a subcomponent answer generation function that resolves $Q_n$ based on retrieved documents $D_n$.

\subsection{Final Answer Generation}
\label{method:Final}
Once each node $n$ in the syntax tree has resolved its local uncertainty with a set of evidence $E_{n}$, \ours{} advances to the final synthesis stage.  To produce the final answer $A$, we prompt the LLM with the full set of node-level evidence $\left\{ E_{n}| n \in N \right\}$ and original question $Q$. The model is guided to synthesize these into a unified response that addresses uncertainties across all sub-phrases: $A=FAG\left\{ Q, \left\{ E_{n}|n \in N \right\} \right\}$, where $FAG$ is a final answer generation prompting function. This phase is responsible for aggregating the distributed, fine-grained inferences across the entire tree into a coherent answer to the original question $Q$. It enables the detection and reconciliation of inconsistencies or conflicting signals that may arise between different pieces of evidence. These inconsistencies can lead to multiple, potentially conflicting answers. Additionally, it ensures logical coherence across the whole tree, validating that intermediate inferences collectively support a consistent global reasoning path.

\section{Experiments}

\label{sec:Experiment}

\subsection{Experiment Setup}\label{exp:step_up}
\textbf{Datasets.}\indent We assess our method on five knowledge-intensive question answering benchmarks that challenge LLMs with multi-step reasoning and ambiguities. For each dataset, we run experiment on $500$ randomly sampled questions. We analyze three multihop question answering datasets: (1) \textbf{HotpotQA} \cite{yang2018hotpotqa}, which contains questions requiring reasoning over multiple supporting paragraphs; (2) \textbf{2WikiMultiHopQA (2WikiMQA)} \cite{ho-etal-2020-constructing}, which consists of entity-centric questions that necessitate combining information from two or more distinct Wikipedia articles; and (3) \textbf{MuSiQue} \cite{trivedi-etal-2022-musique}, which features complex questions composed from simple single-hop questions. We also evaluate performance on ambiguous question answering using two datasets: (1) \textbf{AmbigDocQA} \cite{lee2024ambigdocs}, which contains questions involving ambiguous mentions that may refer to multiple distinct entities, each associated with a different valid answer; and (2) \textbf{ASQA} \cite{stelmakh-etal-2022-asqa}, which contains questions characterized by various types of multifacetedness. 
\\\\
\textbf{Evaluation Metrics.}\indent We employ COVER-EM \cite{rosset2021pretrain}, which assesses whether the generated answer includes the ground truth answer to evaluate multihop questions and ASQA datasets. Following \cite{stelmakh-etal-2022-asqa}, we also use Disambig-F1 (Dis-F1) to evaluate performance on the ASQA dataset.  For AmbigDocQA, we follow the standard evaluation framework introduced by \citet{lee2024ambigdocs}, using Answer Recall (AR) and Entity Recall (ER) as performance metrics.
\\\\
\textbf{Baselines.}\indent We evaluate \ours{} against a comprehensive suite of baselines that represent key prompting and planning paradigms under a unified retriever backbone (BM25 \cite{robertson2009probabilistic}). (1) \textbf{Zero-shot and few-shot prompting} is introduced by \citet{NEURIPS2020_1457c0d6}. These serve as foundational setups without any intermediate reasoning steps. (2) \textbf{Chain-of-Thought prompting} (CoT) \cite{wei2022chain} encourages step-by-step reasoning by appending an instruction such as “Let’s think step by step” to the input. (3) \textbf{Self-Consistency} (SC) \cite{wang2022self} samples multiple reasoning paths and selects the final answer via majority voting. (4) \textbf{Tree-of-Thoughts} (ToT) \cite{yao2023tree} explores multiple structured reasoning trajectories using tree-based search and pruning strategies. (5) \textbf{ReAct} \cite{yao2022react} interleaves reasoning and retrieval by prompting the model to decide dynamically when and what to retrieve. We adapt ReAct to use BM25 instead of web-based tools, denoted as ReAct*. (6) \textbf{Topology-of-Question-Decomposition} (ToQD) \cite{li-etal-2025-topology} constructs a topology graph of sub-questions and uses self-verify inference to selectively activate retrieval only when necessary. Details of baselines implementations are included in Appendix~\ref{app:baseline}.
\\\\
\textbf{Implementation details.}\indent We conduct experiments on three backbone LLMs: GPT-4o-mini \cite{openai2024gpt4omini}, LLaMA3.3-70B \cite{meta2024llama33}, and DeepSeek-V3 \cite{liu2024deepseek}. We implement two variants of \ours{} by adopting two different syntactic formalisms: \ours{} (DT), which leverages dependency trees \cite{culotta-sorensen-2004-dependency} to capture head-dependent relations, and \ours{} (CT), which utilizes constituency trees \cite{langacker1997constituency} to reflect hierarchical phrase structures. We use the Stanza \cite{qi2020stanza} toolkit's dependency and constituency parsers to obtain the required syntactic representations for our framework. For each term-specific query, we employ BM25 \cite{robertson2009probabilistic} to retrieve the top fifteen relevant paragraphs. More specific details of \ours{} are included in Appendix~\ref{app:tree_act}.
\begin{table*}[t]
\small
\renewcommand{\arraystretch}{0.8}
\centering
\setlength{\tabcolsep}{1.5mm}{
\begin{tabular}{llcccc|cc|cc}
\toprule
\multirow{2}{*}{\textbf{Model}}  & \multirow{2}{*}{\textbf{Method}}  & \textbf{HotpotQA} & \textbf{MuSiQue} & \textbf{2WikiMQA} & \multirow{2}{*}{\textbf{AVG}}
& \multicolumn{2}{c|}{\textbf{AmbigDoc}} 
& \multicolumn{2}{c}{\textbf{ASQA}} \\
& & COV-EM & COV-EM & COV-EM & & AR & ER & Dis-F1 & COV-EM \\
\midrule
\multirow{12}{*}{\textbf{GPT4o-mini}} 
& zero-shot & 0.459 & 0.146 & 0.493 & 0.366 & 0.472 & 0.601 & 0.319 & 0.370 \\
& few-shot & 0.473 & 0.151 & 0.520 & 0.381 & 0.409 & 0.539 & 0.328 & 0.423 \\
& COT (zero-shot) & 0.466 & 0.124 & 0.432 & 0.340 & 0.349 & 0.447 & 0.327 & 0.386 \\
& COT (few-shot) & 0.482 & 0.144 & 0.454 & 0.360 & 0.373 & 0.497 & 0.328 & 0.425 \\
&SC (zero-shot) & 0.502 & 0.141 & 0.546 & 0.396 & 0.375 & 0.486 & 0.282 & 0.349 \\
& SC (few-shot) & 0.484 & 0.144 & 0.437 & 0.355 & 0.407 &  0.542 & 0.314 & 0.391 \\
& ReAct & 0.454 & 0.208 & 0.574 & 0.412 & 0.359 & 0.496 & 0.286 & 0.331 \\
& ReAct* & 0.461 & 0.196 & 0.550 & 0.402 & 0.359 & 0.496 & 0.293 & 0.335 \\
& TOT & 0.491 & 0.205 & \textbf{0.612} & 0.436 & 0.187 & 0.535 & 0.255 & 0.265 \\
& ToQD &0.518 &0.188  &0.546    &0.417 & 0.165 &0.182 &0.264  & 0.398 \\
& \cellcolor{lightgray}\textbf{\ours{} (DT)} 
& \cellcolor{lightgray}\textbf{0.544} 
& \cellcolor{lightgray}\underline{0.240} 
& \cellcolor{lightgray}0.583 
& \cellcolor{lightgray}\underline{0.457} 
& \cellcolor{lightgray}\textbf{0.592} 
& \cellcolor{lightgray}\textbf{0.722} 
& \cellcolor{lightgray}\textbf{0.381} 
& \cellcolor{lightgray}\underline{0.547} \\

& \cellcolor{lightgray}\textbf{\ours{} (CT)} 
& \cellcolor{lightgray}\underline{0.542} 
& \cellcolor{lightgray}\textbf{0.264} 
& \cellcolor{lightgray}\underline{0.600} 
& \cellcolor{lightgray}\textbf{0.468} 
& \cellcolor{lightgray}\underline{0.545} 
& \cellcolor{lightgray}\underline{0.642} 
& \cellcolor{lightgray}\underline{0.369} 
& \cellcolor{lightgray}\textbf{0.565} \\
\cmidrule(lr){2-10}
& Rel. Impr.
& 0.083 
& 0.269 
& -0.019
& 0.073 
& 0.254 
& 0.201 
& 0.165 
& 0.329\\
\midrule
\multirow{12}{*}{\textbf{Llama3.3-70B}} 
& zero-shot & 0.516 & 0.134 & 0.521 & 0.390 & 0.476 & 0.558 & 0.344 & 0.478 \\
& few-shot & 0.502 & 0.142 & 0.575 & 0.406 & 0.522 & 0.603 & 0.331 & 0.407 \\
&  COT (zero-shot) & 0.468 & 0.152 & 0.352 & 0.324 & 0.404 & 0.512 & 0.303 & 0.364 \\
& COT (few-shot) & 0.508 & 0.164 & 0.478 & 0.383 & 0.419 & 0.635 & 0.336 & 0.410 \\
& SC (zero-shot) & 0.530 & 0.172 & 0.548 & 0.417 & 0.417 & 0.538 & \underline{0.345} & 0.459 \\
&  SC (few-shot) & 0.532 & 0.168 & 0.563 & 0.421 & 0.411 & 0.586 & 0.331 & 0.384 \\
& ReAct & 0.460 & 0.200 & 0.570 & 0.410 & 0.231 & 0.305 & 0.269 & 0.329 \\
& ReAct* & 0.440 & 0.190 & 0.540 & 0.390 & 0.258 & 0.405 & 0.296 & 0.385 \\
& TOT & 0.404 & 0.195 & \underline{0.603} & 0.401 & 0.185 & 0.530 & 0.283 & 0.423 \\
& ToQD &0.415 &0.202  &0.459    &0.358 &0.189 &0.229 &0.296  &0.355  \\
& \cellcolor{lightgray}\textbf{\ours{} (DT)} 
& \cellcolor{lightgray}\textbf{0.568} 
& \cellcolor{lightgray}\textbf{0.286} 
& \cellcolor{lightgray}\textbf{0.634} 
& \cellcolor{lightgray}\textbf{0.496} 
& \cellcolor{lightgray}\textbf{0.587} 
& \cellcolor{lightgray}\underline{0.686} 
& \cellcolor{lightgray}0.341 
& \cellcolor{lightgray}\textbf{0.518} \\

& \cellcolor{lightgray}\textbf{\ours{} (CT)} 
& \cellcolor{lightgray}\underline{0.540} 
& \cellcolor{lightgray}\underline{0.244} 
& \cellcolor{lightgray}0.584 
& \cellcolor{lightgray}\underline{0.456} 
& \cellcolor{lightgray}\underline{0.568} 
& \cellcolor{lightgray}\textbf{0.704} 
& \cellcolor{lightgray}\textbf{0.357} 
& \cellcolor{lightgray}0.517 \\
\cmidrule(lr){2-10}
& Rel. Impr. 
& 0.068 
& 0.430 
& 0.051
& 0.178
& 0.125 
& 0.165 
& 0.038 
& 0.084\\
\midrule
\multirow{12}{*}{\textbf{Deepseek-V3}} 
& zero-shot & 0.512 & 0.146 & 0.547 & 0.401 & 0.521 & 0.632 & 0.348 & 0.451 \\
& few-shot & 0.526 & 0.154 & 0.550 & 0.410 & 0.545 & 0.661 & 0.358 & 0.411 \\
& COT (zero-shot) & 0.498 & 0.142 & 0.426 & 0.355 & 0.405 & 0.527 & 0.347 & 0.416 \\
&  COT (few-shot) & 0.533 & 0.163 & 0.526 & 0.395 & 0.426 & 0.599 & 0.358 & 0.419 \\
& SC (zero-shot) & 0.513 & 0.146 & 0.574 & 0.411 & 0.446 & 0.631 & 0.362 & 0.465 \\
& SC (few-shot) & 0.524 & 0.152 & 0.574 & 0.417 & 0.421 & 0.578 & 0.352 & 0.448 \\
& ReAct & 0.503 & 0.252 & \underline{0.673} & 0.476 & 0.266 & 0.352 & 0.283 & 0.308 \\
& ReAct* & 0.479 & 0.264 & 0.630 & 0.458 & 0.284 & 0.371 & 0.308 & 0.477 \\
& TOT & 0.505 & 0.273 & 0.551 & 0.443 & 0.174 & 0.513 & 0.214 & 0.308 \\
& ToQD &0.448 &0.216  &0.451    &0.371 &0.215 &0.275 &0.327  &0.408  \\
& \cellcolor{lightgray}\textbf{\ours{} (DT)} 
& \cellcolor{lightgray}\underline{0.572} 
& \cellcolor{lightgray}\textbf{0.280} 
& \cellcolor{lightgray}0.650 
& \cellcolor{lightgray}\underline{0.501} 
& \cellcolor{lightgray}\underline{0.567} 
& \cellcolor{lightgray}\underline{0.667} 
& \cellcolor{lightgray}\textbf{0.406} 
& \cellcolor{lightgray}\underline{0.558} \\
& \cellcolor{lightgray}\textbf{\ours{} (CT)} 
& \cellcolor{lightgray}\textbf{0.594} 
& \cellcolor{lightgray}\underline{0.278} 
& \cellcolor{lightgray}\textbf{0.674} 
& \cellcolor{lightgray}\textbf{0.515} 
& \cellcolor{lightgray}\textbf{0.589} 
& \cellcolor{lightgray}\textbf{0.721} 
& \cellcolor{lightgray}\underline{0.391} 
& \cellcolor{lightgray}\textbf{0.566} \\
\cmidrule(lr){2-10}
& Rel. Impr. 
& 0.114 
& 0.026 
& 0.001
& 0.082
& 0.131 
& 0.091 
& 0.122 
& 0.187\\
\bottomrule
\end{tabular}
}
\caption{Performance of \ours{} (CT) \ours{}(DT), Chain-of-Thought (CoT), Self-Consistency (CoT-SC), ReAct, ReAct*, Tree-of-Thoughts (ToT) and Topology-of-Question-Decomposition (ToQD) across five different QA datasets. AVG indicates the average COV-EM on the three multihop datasets. ReAct* denotes a BM25-based variant of ReAct. \textbf{Bold} marks the best performance, and \underline{underline} denotes the second-best under teh same setting. Rel. Impr. stands for relative improvement over the best baseline in the same setting. }
\label{main-results}
\end{table*}

\subsection{Results}\label{exp:result}
Table~\ref{main-results} presents the evaluation results of \ours{} against competitive baselines across three LLM backbones. In general, \ours{} leads or closely matches top-performing baselines in both multihop and ambiguous QA tasks, demonstrating the effectiveness of \ours{} in handling complex, knowledge-intensive QA. 
\begin{table*}[t]
\small
\centering
\setlength{\tabcolsep}{1.5mm}{
\begin{tabular}{lcccc|cc|cc}
\toprule
\multirow{2}{*}{\textbf{Retriever}} & \textbf{HotpotQA} & \textbf{MuSiQue} & \textbf{2WikiMQA} & \multirow{2}{*}{\textbf{AVG}} 
& \multicolumn{2}{c|}{\textbf{AmbigDoc}} 
& \multicolumn{2}{c}{\textbf{ASQA}} \\
& COV-EM & COV-EM & COV-EM &
& AR & ER 
&Dis-F1& COV-EM   \\
\midrule
BM25 & 0.473 & 0.151 & 0.520 & 0.381 & 0.409 & 0.539 & 0.328 & 0.423 \\
DPR & 0.392 & 0.132 & 0.318 & 0.281 & 0.343 & 0.449 & 0.343 &  0.438\\
BM25+DPR & 0.456 & 0.146 & 0.380 & 0.327 & 0.322 & 0.404 & 0.351 &0.480  \\
IRCoT & 0.486 & 0.148 & 0.504 & 0.379 & 0.314 & 0.394 &  0.335 &0.488
\\
Tree-Retrieval (DT) & \textbf{0.528} & 0.138 & \textbf{0.562} & \textbf{0.409} & 0.427 & 0.581 &\textbf{0.361}  & 0.514 \\
Tree-Retrieval (CT) & 0.510 & \textbf{0.156} & 0.538 & 0.401 & \textbf{0.558} & \textbf{0.681} & 0.335  & \textbf{0.533} \\
\bottomrule
\end{tabular}
}
\caption{Performance comparison of Tree-Retrieval and standard retrievers on multihop (HotpotQA, MuSiQue, 2WikiMQA) and ambiguous QA datasets (AmbigDoc, ASQA). The AVG column represents the mean COV-EM across multihop datasets.}
\label{tab:retriever_comparison}
\end{table*}

In multihop QA, \ours{} achieves the best or near-best scores in most cases. \ours{} (CT) under DeepSeek-V3 achieves the strongest performance, with an average COVER-EM of $0.515$ on multihop QA and a relative improvement of $0.082$ over the best-performing baseline. Similar trends are observed under LLaMA3.3-70B and GPT4o-mini. The only exception is 2WikiMQA under GPT4o-mini, where TOT slightly outperforms \ours{}, likely because the extended evidence derived from the tree structure exceeds GPT4o-mini’s limited reasoning capacity. ReAct performs well with stronger backbones but degrades under smaller models, reflecting its reliance on effective prompt-based reasoning. In contrast, \ours{} demonstrates robust performance across different model scales.


\ours{} shows remarkable gains on AmbigDoc and ASQA across all backbones. In particular, under DeepSeek-V3, \ours{} (DT) achieves the highest Dis-F1 score of $0.406$ and the largest relative improvement in COVER-EM in ASQA. These results demonstrate that grounding retrieval in each sub-phrase and subsequently aggregating the collected information helps uncover signals of different plausible interpretations. Notably, on AmbigDoc, \ours{} (CT) also achieves the highest AR score of $0.592$, outperforming the best baseline by a substantial margin. This highlights \ours{}'s strength in disambiguating entities that share the same name.


\section{Analysis}

\subsection{Tree-Retrieval}
To demonstrate the contribution of syntactic decomposition to retrieval quality, we devise Tree-Retrieval that mirrors \ours{} while eliminating all the module that require LLM's reasoning. Specifically, we discard the subcomponent-based query generation module and instead directly utilize the corresponding sub-phrases to retrieve relevant documents from the corpus. Then we employ a reranking model to select the top 15 passages across the sub tree rooted at each node. These top passages serve as a substitute for the evidence generated by subcomponent question answering in \ours{}. Reranking model evaluates the initially retrieved documents using more sophisticated models to better assess their relevance to the subcomponent \cite{kratzwald-etal-2019-rankqa}. Implementation details and experiments on other backbone models are provided at Appendix~\ref{app:tree}.

As shown in Table~\ref{tab:retriever_comparison}, Tree-Retrieval consistently surpasses classical retrieval approaches such as BM25, DPR\cite{karpukhin-etal-2020-dense}, and LLM based question decomposition approaches like IRCoT \cite{trivedi-etal-2023-interleaving} across all QA datasets. The performance gain shows that incorporating syntactic structure into the retrieval process enhances the relevance of retrieved documents. Additionally, these results suggest that the effectiveness of \ours{} cannot be solely attributed to enhanced downstream reasoning by the LLMs. Rather, a significant portion of its advantage stems from its retrieval stage, which is structurally guided to extract more fine-grained, contextually aligned documents. 
\begin{table*}[t]
\small
\centering
\setlength{\tabcolsep}{1.5mm}{
\begin{tabular}{lcccc|cc|cc}
\toprule
\multirow{2}{*}{\textbf{Ablation}}  & \textbf{HotpotQA} & \textbf{MuSiQue} & \textbf{2WikiMQA} &\multirow{2}{*}{\textbf{AVG}}
& \multicolumn{2}{c|}{\textbf{AmbigDoc}} 
& \multicolumn{2}{c}{\textbf{ASQA}} 
 \\
 &COV-EM & COV-EM & COV-EM &
& AR & ER 
& COV-EM & Dis-F1 \\
\midrule
\ours{} (DT) \textit{w/o} QG&0.514 &0.214 &0.537 &0.422 &0.624 &0.797 &0.337 &0.454\\
\ours{} (DT) \textit{w/o} SAG&0.514 &0.155 &0.472 &0.380 &0.589 &0.766 &0.360 &0.485\\
\ours{} (DT) \textit{w/o} IR&0.431 &0.156 &0.366 &0.318 &0.244 &0.262 & 0.285&0.403\\
\midrule
IR Only &0.450 & 0.146 & 0.490 & 0.362 & 0.472 & 0.601 & 0.319 & 0.371 \\
QG Only &0.352 &0.124 &0.448 &0.308 &0.183 &0.197& 0.248 &0.350 \\
COT Only &0.373&	0.146&	0.334	&0.284&	0.172	&0.186 &0.284&	0.333\\
\bottomrule
\end{tabular}
}
\caption{Ablation study on \ours{} based on Dependency Tree where the backbone model is GPT4o-mini. "QG","SAG","IR" refer to the subcomponent queries generation, subcomponent answer generation and information retrieval. }
\label{ableviation_study}
\end{table*}

\subsection{Ablation studies}
We conduct a series of systematic ablation experiments to evaluate the importance of each single module in \ours{}. Specifically, we evaluate the impact of (1) subcomponent-based query generation (QG), (2) information retrieval (IR), and (3) subcomponent question answering (SAG). In each setting, we selectively disable one module, and we additionally assess configurations where only a single module is retained. Implementation details of these ablation experiments are provided in the Appendix~\ref{app:abla}.  

Table~\ref{ableviation_study} shows that removing any core module from \ours{} leads to a significant performance drop, confirming their complementary importance. First, the largest decreases are observed in removal of information retrieval, indicating that information retrieval is the most essential component for generating the correct answer. Second, the absence of question generation causes moderate performance drop (average COVER-EM down to $0.419$). Therefore, subcomponent-based queries enhance retrieval relevance and correctness of each reasoning step. Third, removing subcomponent question answering results in substantial degradation, with average COVER-EM decreasing to $0.318$. This finding aligns with the ``Lost-in-the-Middle'' \cite{liu-etal-2024-lost} phenomenon. Interestingly, its removal subtly improves performance on AmbigDoc, suggesting that entity-specific cues might be lost during LLM-based filtering.

Among the single-module configurations, the IR-only setting achieves the highest overall performance, confirming the dominant role of information retrieval in knowledge-intensive QA. In contrast, QG-only and COT-only variants perform poorly. As QG-only generates target queries, it outperforms COT-only on ambiguous QA.
\begin{figure}[t]
    \centering
    \includegraphics[width=1\linewidth]{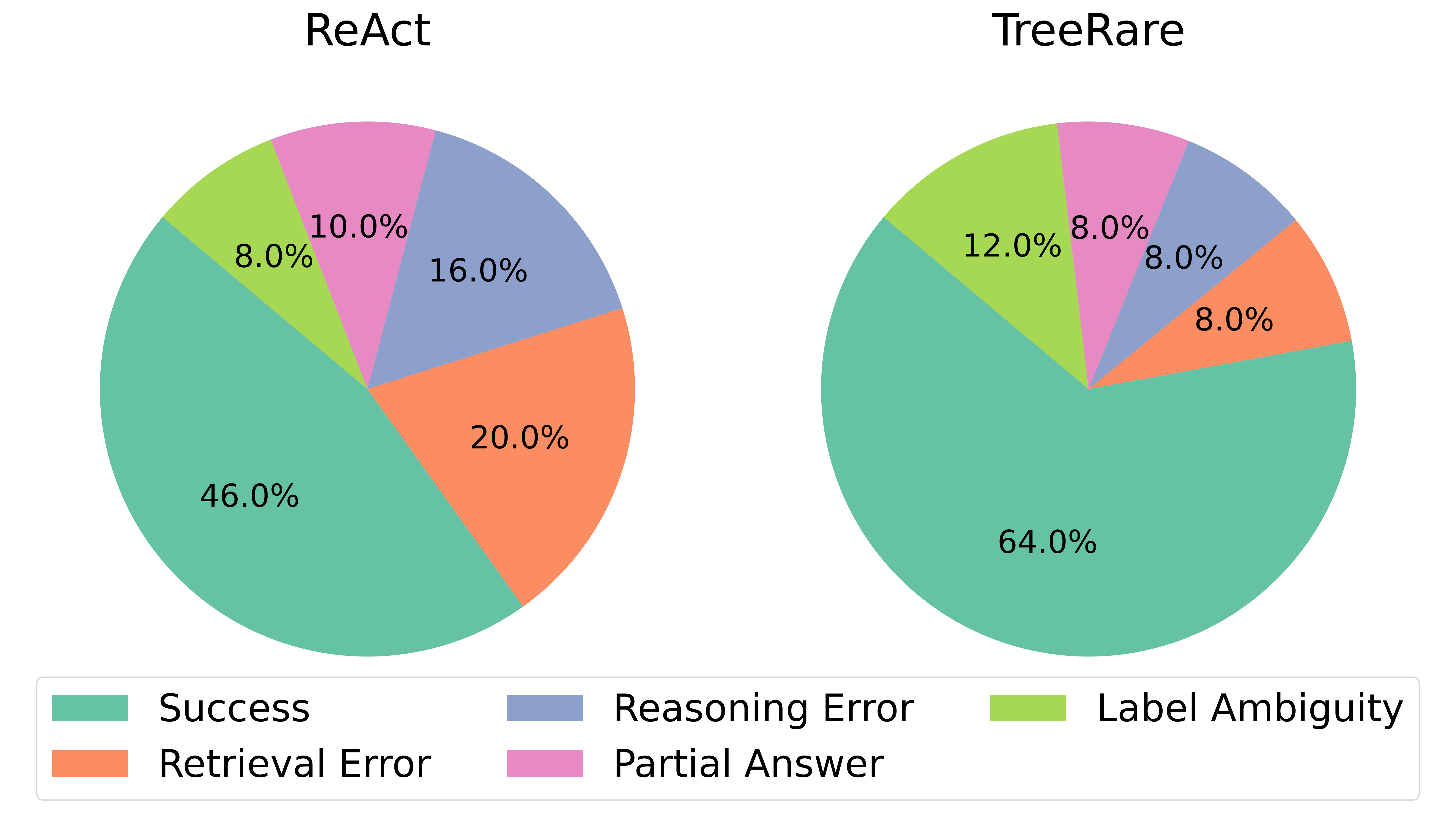}
    \caption{Distribution of outcome types for ReAct and \ours{} on randomly sampled multihop QA.  }
    \label{fig:case_study}
\end{figure}
\subsection{Error Analysis}\label{abla:case study}
To better understand the performance differences between ReAct and \ours{}, we conduct a human evaluation comparing their outputs on randomly sampled multihop questions. Each output is manually categorized into one of five distinct outcome types: \textit{Success}, \textit{Retrieval Error}, \textit{Reasoning Error}, \textit{Partial Answer}, and \textit{Label Ambiguity}. The detailed definitions and illustrative examples of each error type are presented in Appendix~\ref{app:case_study}.

As shown in Table~\ref{fig:case_study}, \ours{} achievees a higher success rate compared to ReAct\cite{yao2022react}, indicating \ours{}'s effectiveness in enhancing the LLM's ability to produce correct answers. ReAct exhibits a high rate of retrieval errors, suggesting a lack of effective guidance in query generation. This is primarily due to its heavy reliance on few-shot prompting and the model's reasoning abilities. In contrast, \ours{} offers explicit structural guidance through syntax tree, which results in more effective queries to guide retrieval and a large decrease in the retrieval error rate. Additionally, \ours{} reduces the rate of partial answer by $2\%$ and reasoning errors by $8\%$, suggesting that its structured guidance mechanism better supports the reasoning alignment. Meanwhile, \ours{} has a higher proportion of label ambiguity. While reflecting a higher incidence of mismatch with labeled answers, \ours{} may in fact produce correct responses that differ from annotated references.

\subsection{Cost Analysis}
\begin{figure}[t]
    \centering
    \includegraphics[width=1\linewidth]{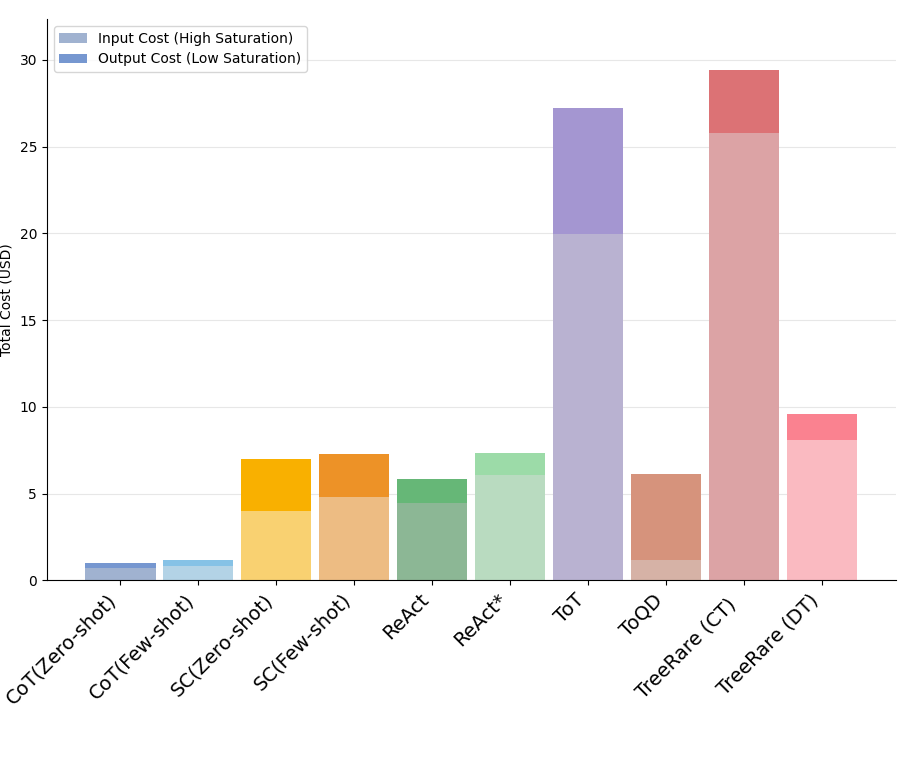}
    \caption{Total GPT-4o-mini API cost for \ours{} (CT), \ours{} (DT), TOT, ReAct, SC, and COT across HotpotQA, MuSiQue, 2WikiMQA, AmbigDoc, and ASQA}
    \label{fig:cost_comparison}
\end{figure}

To assess the computational efficiency of \ours{}, we measure the total number of input and output tokens generated during inference on $500$ randomly sampled examples per dataset. The token usage is translated into cost according to the GPT-4o-mini pricing scheme published by OpenAI. Figure~\ref{fig:cost_comparison} illustrates the cost breakdown across different methods. We observe that \ours{} (CT) incurs higher inference cost compared to \ours{} (DT). This difference can be attributed to the structural characteristics of constituency trees, which represent nested phrase structures and thus tend to include more nodes and sub-phrases than dependency trees. Since \ours{} performs query generation and retrieval at each node, deeper trees with more branches lead to increased token usage. Furthermore, since  methods such as ToT and SC involve extensive sampling of reasoning trajectories, they inflate both input and output token counts and exhibit significantly high computational costs. \ours{} (DT) offers a favorable trade-off, achieving better performance with moderate computational demands. 

\section{Conclusion}
In conclusion, we propose \ours{} for knowledge-intensive question answering, utilizing syntax trees to guide information retrieval and structural reasoning. When traversing the syntax tree, our method performs subcomponent-based information retrieval and question answering. This structured approach enhances retrieval quality and models' ability to resolve information gaps at each node. Experimental results across knowledge-intensive benchmarks demonstrate that our method achieve significant performance improvements over state-of-the-art baselines.
\section*{Limitation}
While \ours{} demonstrates strong performance across multihop and ambiguous question answering datasets, several limitations remain.

First, \ours{} relies on the quality and granularity of syntax parsers. Errors in dependency or constituency parsing may propagate through the bottom-up reasoning pipeline, leading to suboptimal subcomponent decomposition and misaligned query formulation.

Second, \ours{} incurs additional computational overhead due to its multi-stage decomposition, retrieval, and filtering pipeline. This overhead is particularly pronounced for the constituency tree variant, which typically produces deeper and more richly branched trees, resulting in increased token usage. Such cost implications may hinder \ours{}’s scalability in latency- or budget-constrained deployment settings.

Third, \ours{} has been evaluated exclusively on factoid-style questions, where each query maps to discrete factual answers. Its performance on open-domain dialogue or generative settings—such as those requiring opinion modeling, pragmatic reasoning, or user intent tracking—remains unexplored. 


\bibliography{anthology,custom}

\appendix
\clearpage
\section{\ours{} Implementation Details}\label{app:tree_act}
\textbf{Parsing Module.} To construct the syntax trees required by \ours{}, we utilize the dependency parser and constituency parser from Stanza\cite{qi2020stanza}. For each input question, we parse it into both dependency and constituency structures. We implement a unified interface to map parsed trees into a bottom-up traversal format, ensuring that each node contains: (i) its corresponding sub-phrase span, (ii) its child nodes. (iii) syntactic type (e.g., NP, VP for constituency, or head-dependent relation for dependency).\\\\
\textbf{Traversing Details} In the experiment, \ours{} conducts a pre-order traversal. Each node is processed only after all its child nodes are resolved. We maintain a processing queue initialized with leaf nodes. When processing a node: If it has no children, we directly use its text span to generate quries. If it has children, we first aggregate evidence from its children before proceeding to subcomponent-specific query generation. If it has no children, we directly use its text span as the initial evidence. To improve computational efficiency, we implement a pruning mechanism that skips nodes based on two criteria: Nodes with syntactic types typically considered non-informative (e.g., punctuation, determiners, conjunctions). Nodes whose associated sub-phrases are minimum phrase length ($L_{min}$). We sets $L_{min}=3$ for all the experiments in the paper. In\\\\
\textbf{Subcomponent Query Generation.} At each non-leaf node, we generate queries to resolve syntactic uncertainty associated with its sub-phrase. We prompt the LLM using templates detailed in Figure\ref{prompt:A}, Figure\ref{prompt:B}, Figure\ref{prompt:C}, Figure\ref{prompt:D}. We generate up to five candidate queries and select the top three queries according to heuristic rules prioritizing coverage and specificity.  \\\\
\textbf{Retrieval Module.} We use BM25 via Pyserini \cite{Lin_etal_SIGIR2021_Pyserini} as the retrieval backend. For each generated query, we retrieve the top-15 paragraphs  from Wikipedia dump in \cite{karpukhin-etal-2020-dense}. If multiple queries exist for a node, their retrieved documents are merged. In Table~\ref{tab:compar tree}, we present the comparison of the performance between BM25 and DPR\cite{karpukhin-etal-2020-dense} on \ours{}. We observe that BM25 outperforms DPR under five benchmark, and thus we implement \ours{} with BM25 as the backbone retriever. 

\section{Baseline Implementation Details}\label{app:baseline}
To ensure a fair comparison with \ours{}, we implemented all baseline prompting methods within a direct RAG setup using a shared retrieval backbone. Specifically, we employ BM25 \cite{robertson2009probabilistic} as the sparse retriever and retrieve the top-20 most relevant passages from a Wikipedia corpus for each query. We directly use the implementation from Pyserini \cite{Lin_etal_SIGIR2021_Pyserini}. The retrieved contexts are concatenated with the input prompt and passed to the large language model, leveraging its extended context window. 

\begin{table*}[t]
\small
\centering
\setlength{\tabcolsep}{1.5mm}{
\begin{tabular}{lcccc|cc|cc}
\toprule
\multirow{2}{*}{\textbf{Retriever}} & \textbf{HotpotQA} & \textbf{MuSiQue} & \textbf{2WikiMQA} & \multirow{2}{*}{\textbf{AVG}} 
& \multicolumn{2}{c|}{\textbf{AmbigDoc}} 
& \multicolumn{2}{c}{\textbf{ASQA}} \\
& COV-EM & COV-EM & COV-EM &
& AR & ER 
&COV-EM & Dis-F1  \\
\midrule
Tree-Retrieval (DT) \textit{w} BM25 & 0.528 & 0.138 & 0.562 & 0.409 & 0.427 & 0.581 &0.514  & 0.361 \\
Tree-Retrieval (DT) \textit{w} DPR & 0.482& 0.16& 0.504 & 0.382 & 0.409 & 0.479 & 0.399 & 0.332   \\
\bottomrule
\end{tabular}
}
\caption{Comparison of Tree-Retrieval (DT) with BM25 and DPR as retriever backbone.}
\label{tab:compar tree}
\end{table*}

\textbf{Few-shot Prompting.} For few-shot prompting, we prepend three in-context examples drawn from the same dataset as the test instance

\textbf{Chain-of-Thought Prompting.} We follow the original CoT formulation \cite{wei2022chain}, appending reasoning demonstrations to the prompt to elicit step-by-step inference. In the few-shot CoT setting, each demonstration consists of a question, a multi-step rationale, and the final answer. In zero-shot CoT, the test query is prefixed by the phrase “Let’s think step by step.”

\textbf{Self-Consistency.} For multihop question, we generate 10 independent reasoning trajectories using the few-shot CoT prompt. The model’s final answer is selected by majority vote among the answers extracted from each reasoning trace. However, SC is not suitable for long-form answer generation. In AmbigDoc and ASQA, we follow the USC\cite{chen2024universal} protocol for self-consistency.

\textbf{React.} We closely follow the original ReAct framework as proposed by \cite{yao2022react}. The maximum number of steps is set to eight. If the model fails to reach a conclusive answer within this limit, we default to using Self-Consistency prompting to generate the final response. For ambiguous questions, we adopt the query refinement prompts introduced by \cite{amplayo-etal-2023-query} abd incorporate few-shot exemplars directly into the prompt.
\begin{figure*}[t]
    \centering
    \includegraphics[width=1\linewidth]{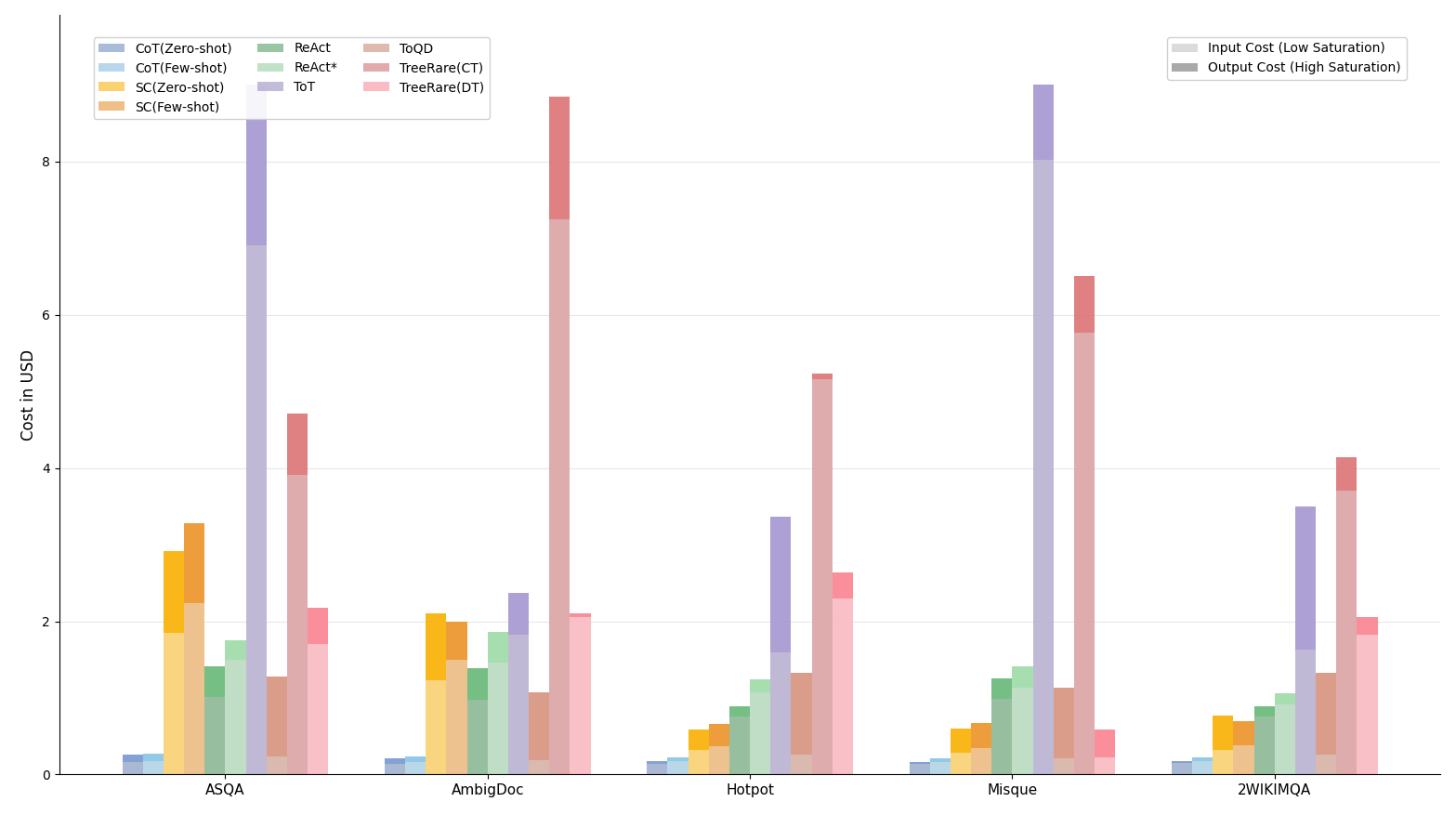}
    \caption{Total GPT-4o-mini API cost (input + output tokens) for \ours{}(CT), \ours{}(DT), TOT, ReAct, SC, and COT across HotpotQA, MuSiQue, 2WikiMQA, AmbigDoc, and ASQA based on OpenAI pricing.}
    \label{fig:total_cost_comparison}
\end{figure*}

\textbf{Tree-of-Thoughts.} The original Tree-of-Thought paper does not provide a pipeline tailored for multihop or ambiguous question answering. Therefore, we implement ToT following the setup in \cite{pmlr-v235-zhou24r}. Instead of sampling multiple reasoning paths as in the original version, our implementation adopts the React framework to sample diverse plan-and-action paths, enabling interaction with Wikipedia and equipping ToT with enhanced capabilities for open-domain QA. 

\textbf{Topology-of-Question-Decomposition.} We follow the implementation of the ToQD paper. To make sure every method uses the same retriever, we modify the retrieval module of ToQD into the same as the retrieval module in \ours{}.  

\section{Ablation Study Details}\label{app:abla}
We do not include a setting with only subcomponent question answering in our ablation study. This is because it is not feasible to generate fine-grained answers without first performing query generation and retrieval.\\\\
\textbf{\ours{}(DT) \textit{w/o} CQ.}\indent To assess the impact of subcomponent-based query generation, we replace the \ours{} query generation module with a naive retrieval. Specifically, for each node in the syntactic tree, we bypass the LLM-generated term-specific queries and instead directly use the surface form of the corresponding sub-phrase as the retrieval query. In the downstream subcomponent answering stage, we prompt LLMs to directly answer the question with the retrieved documents.\\\\
\textbf{\ours{}(DT) \textit{w/o} QA.} \indent In this variant, we eliminate the intermediate reasoning step that resolves each node’s syntactic uncertainty. Instead of prompting the LLM to process the retrieved evidence at each node, we directly aggregate all retrieved documents across sub-nodes and forward the combined evidence to their parent node without further interpretation.

\begin{table*}[t]
\small
\centering
\setlength{\tabcolsep}{1.5mm}{
\begin{tabular}{lcccc|cc|cc}
\toprule
\multirow{2}{*}{\textbf{$L_{min}$}}  & \textbf{HotpotQA} & \textbf{MuSiQue} & \textbf{2WikiMQA} &\multirow{2}{*}{\textbf{AVG}}
& \multicolumn{2}{c|}{\textbf{AmbigDoc}} 
& \multicolumn{2}{c}{\textbf{ASQA}} 
 \\
 &COV-EM & COV-EM & COV-EM &
& AR & ER 
& COV-EM & Dis-F1 \\
\midrule
3& 0.544& 0.240& 0.583 & 0.457& 0.592& 0.722& 0.381&0.547\\
6&0.531 &0.224 &0.547  &0.434 & 0.542&0.703 &0.333 &0.441\\
10&0.512 &0.198 &0.492&0.318 &0.504 &0.600 & 0.328&0.448\\
\bottomrule
\end{tabular}
}
\caption{Impact of minimum phrase length ($L_{min}$) on \ours{} (DT) performance across across HotpotQA, MuSiQue, 2WikiMQA, AmbigDoc, and ASQA based on OpenAI pricing.}
\label{Table:miminum}
\end{table*}
\section{Tree Retrieval}\label{app:tree}
For each node in the syntax tree, we directly use sub-phrase $p_n$ without any LLM-based reformulation. This phrase is used as a query to the retrieval systems. We retrieve the top10 passages from Wikipedia corpus using BM25. The retrieved passages across all nodes within a sub-tree are merged. To suppress noise and prioritize passages most relevant to the sub-tree's syntactic content, we apply a cross-encoder reranker MS-Marco-MiniLM-L-12-v2. In Table~\ref{tab:compar tree}, we compared Tree-Retrieval with different both BM25 and DPR\cite{karpukhin-etal-2020-dense}. 
\section{Further Cost Analysis}
As shown in Figure\ref{fig:total_cost_comparison}, we measure the API cost of GPT-4o-mini by summing the number of input and output tokens processed for each method across five datasets: HotpotQA, MuSiQue, 2WikiMultihopQA, AmbigDoc, and ASQA. For each dataset, 500 examples were randomly sampled and processed with each method. The input and output tokens were multiplied by OpenAI’s published pricing for GPT-4o-mini (as of April 2025).

\section{Further Analysis of Minimum Phrase Length}\label{app:minimum_phrase_Length}
\begin{figure*}[t]
    \centering
    \includegraphics[width=1\linewidth]{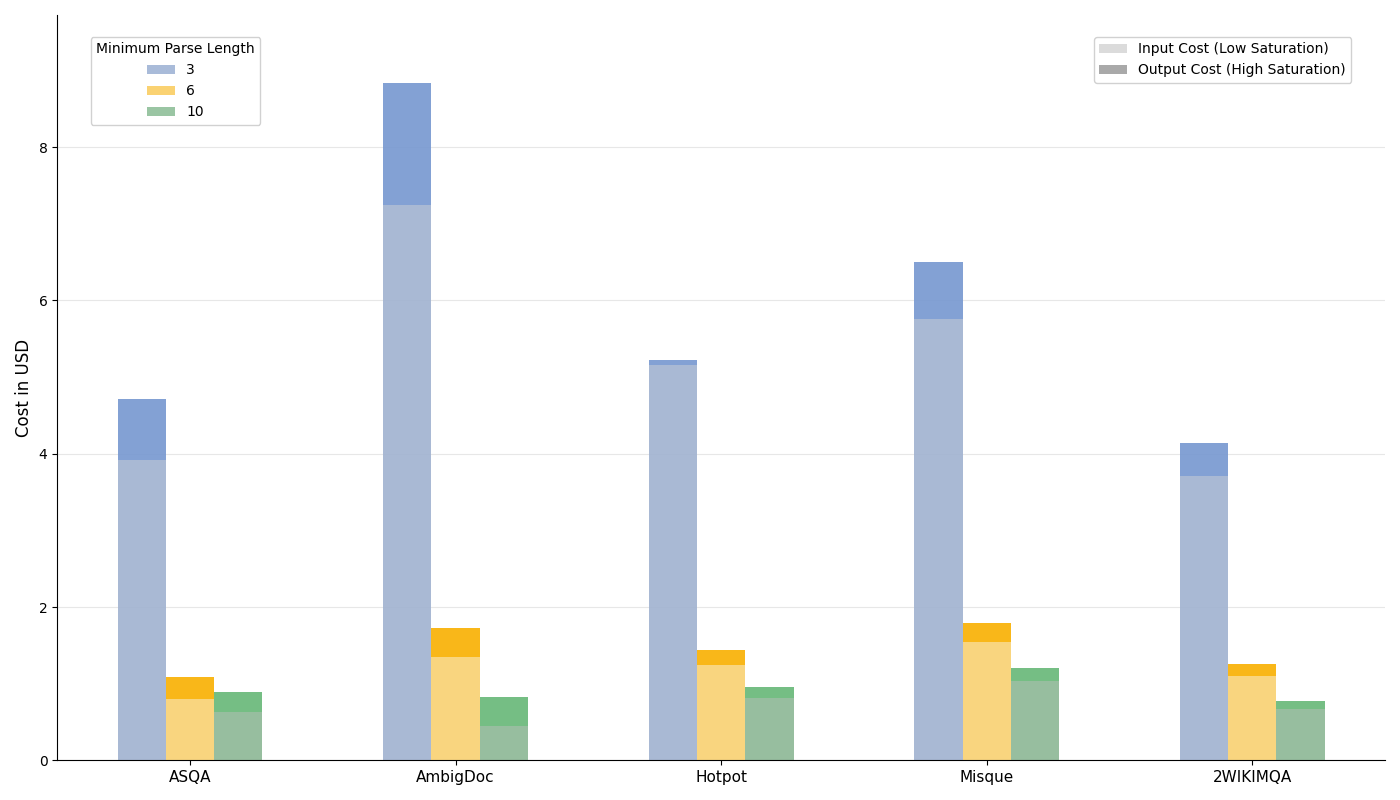}
    \caption{Total GPT-4o-mini API cost (input + output tokens) for \ours{}(CT) on different minimum phrase lengths (3,6,10) across HotpotQA, MuSiQue, 2WikiMQA, AmbigDoc, and ASQA based on OpenAI pricing.}
    \label{fig:minimum_phrase_length}
\end{figure*}

To analysis the impact of minimum phrase length of \ours{}, we performs additional experiment by setting $L_{min}$ to $6$ and $10$. As shown in \ref{Table:miminum}, Table\ours{} demonstrates consistently superior performance with $L_{min}=3$ across most evaluation metrics. The performance degradation with increased $L_{min}$ values suggests that finer-grained syntactic analysis enables more precise identification of knowledge gaps and uncertainty points within complex questions. Figure \ref{fig:minimum_phrase_length} reveals the computational trade-offs associated with different $L_{min}$. $L_{min}$ values result in higher API costs due to increased node processing. This cost increase stems from processing more nodes in the syntax tree, each requiring query generation, retrieval, and subcomponent question answering.

\section{Case Study}\label{app:case_study}
Here is the definition for different categories: 
\begin{itemize}
    \item \textbf{Success}: Model produces a correct and complete answer aligned with the reference.
    \item \textbf{Retrieve Error}: Retrieved documents are irrelevant or do not aid answer resolution.
    \item \textbf{Reasoning Error}: Model performs faulty reasoning despite relevant evidence being available.
    \item \textbf{Partial Answer}: Answer is generally correct but lacks necessary specificity.
    \item \textbf{Label Ambiguity}: Prediction is plausible but does not exactly match the labeled answer.
\end{itemize}

we randomly sampled 50 multihop questions from the HotpotQA development set. For each sample, a human annotator examined the reasoning trace produced by both ReAct and \ours{}, along with the final answer and supporting documents. We present one representative example for each failure category. Additionally, we provide a single case study of concrete output \ours{} on one single question at Figure\ref{output1} and Figure\ref{output2}. 
\begin{figure*}
\begin{tcolorbox}[colback=white!5!white,colframe=blue!75!black,title=Label Ambiguity]
\textbf{\ours{}:}\\
\textbf{Question:} Roger O. Egeberg was Assistant Secretary for Health and Scientific Affairs during the administration of a president that served during what years?\\
.......\\
\textbf{Final answer:} Roger O. Egeberg was Assistant Secretary for Health and Scientific Affairs during the Nixon administration, which lasted \textcolor{red}{from January 20, 1969, to August 9, 1974}. (Label: 1969 until 1974)\\
\textbf{ReAct:}\\
\textbf{Question:} Which other Mexican Formula One race car driver has held the podium besides the Force India driver born in 1990?\\
.......\\
\textbf{Final answer:} \textcolor{red}{Pedro Rodríguez de la Vega} (Label: Pedro Rodríguez)\\
\end{tcolorbox}
\caption{Examples of \textit{Label Ambiguity}. The ambiguous span is highlighted in \textcolor{red}{red}.}
\end{figure*}
\begin{figure*}
\begin{tcolorbox}[colback=white!5!white,colframe=blue!75!black,title=Partial Answer]
\textbf{\ours{}:}\\
\textbf{Question:} What is the name of the fight song of the university whose main campus is in Lawrence, Kansas and whose branch campuses are in the Kansas City metropolitan area?\\
.......\\
\textbf{Final answer:}
The fight song of the University of Kansas is \textcolor{red}{"I'm a Jayhawk."}(Label: Kansas Song)\\
\textbf{ReAct:}\\
\textbf{Question:} The director of the romantic comedy "Big Stone Gap" is based in what New York city?\\
.......\\
\textbf{Final answer:} \textcolor{red}{New York City}\\
\end{tcolorbox}
\caption{Examples of \textit{Partial Answer}. The partial answer is highlighted in \textcolor{red}{red}.}
\end{figure*}
\begin{figure*}
\begin{tcolorbox}[colback=white!5!white,colframe=blue!75!black,title=Search Error]
\textbf{\ours{}:}\\
\textbf{Question:} What type of forum did a former Soviet statesman initiate?
\textbf{Generate Question:} What is the \textcolor{red}{name of the former Soviet statesman} who initiated the forum? ...
\textbf{answer:} The name of the former Soviet statesman who initiated a notable forum is Yegor Gaidar....
\\
\textbf{Generate Question:} What was the name of the former Soviet statesman? ...
\textbf{answer: }The name of the former Soviet statesman is Mikhail Suslov ....
 question: What was the role of Mikhail Suslov in the Soviet Union? ....\\
answer: Mikhail Suslov served as the Second Secretary .....\\
\textbf{Generate Question:} What topics are typically discussed in the forum organized by the former Soviet statesman? ....
\textbf{answer:} The \textcolor{red}{document does not contain specific information} about a political forum initiated by a former Soviet statesman ... \\
\textbf{Fianl answer:}
former Soviet statesman, Yegor Gaidar, initiated the Gaidar Forum, which focuses on discussions among Russia's political and business elite regarding economic reforms and political strategies.
\textbf{ReAct:}\\
\textbf{Question:} This singer of A Rather Blustery Day also voiced what hedgehog?\\
Search[A Rather Blustery Day]
Thought 1: I need to search for the song "A Rather Blustery Day"...\\
Action 1: Search[A Rather Blustery Day]
Observation 1: Winnie the Pooh and the Blustery Day is a 1968 American animated musical fantasy short ...\\
Lookup[singer of A Rather Blustery Day]
Thought 2: \textcolor{red}{The search didn't give me the specific singer ...} \\
Action 2: Lookup[singer of A Rather Blustery Day]
Observation 2: No more results.\\
Search[A Rather Blustery Day singer]
Thought 3: \textcolor{red}{The search did not yield any direct information} \\
Action 3: Search[A Rather Blustery Day singer]
Observation 3: Could not find A Rather Blustery Day singer\\
Search[hedgehog character voiced by a singer from Winnie the Pooh]
Thought 4: \textcolor{red}{The search did not provide specific details} \\
Action 4: Search[hedgehog character voiced by a singer from Winnie the Pooh]
Observation 4: Could not find hedgehog character \\
Search[hedgehog character in animated films]
Thought 5: \textcolor{red}{The search has not provided concrete information} \\
Action 5: Search[hedgehog character in animated films]
Observation 5: Could not find hedgehog character \\
Search[singer of Sonic the Hedgehog]
Thought 6: It seems that I'm stuck \\
Action 6: Search[singer of Sonic the Hedgehog]
Observation 6: Could not find singer of Sonic the 
\end{tcolorbox}
\caption{Examples of \textit{Search Error}. The search result or incorrect query is highlighted in \textcolor{red}{red}.}
\end{figure*}
\begin{figure*}
\begin{tcolorbox}[colback=white!5!white,colframe=blue!75!black,title=Incorrect reasoning]
\textbf{\ours{}:}\\
\textbf{Question:} question: Brown State Fishing Lake is in a country that has a population of how many inhabitants ?\\
\textbf{Generate Question:} What country is Brown State Fishing Lake located in?;
\textbf{answer:} Brown State Fishing Lake is located in the United States .... \\
\textbf{Generate Question:} What is the population of the \textcolor{red}{ United States as of 2023}?
\textbf{answer:}The population of the United States as of 2023 is approximately 333 million people. ....
\\\textbf{Final answer:}
The population of the United States, where Brown State Fishing Lake is located, is approximately 333 million as of 2023. (label: 9,984)\\
\textbf{ReAct:}\\
\textbf{Question:} Which performance act has a higher instrument to person ratio, Badly Drawn Boy or Wolf Alice? 
Search[Badly Drawn Boy]
Thought 1: I need to search both Badly Drawn Boy and Wolf Alice .....
Action 1: Search[Badly Drawn Boy]
Observation 1: Damon Michael Gough (born 2 October 1969), ..... and multi-instrumentalist.
\\
Lookup[Badly Drawn Boy performances]
Thought 2: .... I need to search for more explicit information about his performances.

Action 2: Search[Wolf Alice]
Observation 2: Wolf Alice are an English rock band fr....  acoustic duo comprising singer Ellie Rowsell and guitarist Joff Oddie, Wolf Alice have also featured bassist Theo Ellis and drummer Joel Amey since 2012 ....
Lookup[Wolf Alice instruments]
Thought 4: Wolf Alice consists of four members: Ellie Rowsell (vocals), Joff Oddie (guitar), Theo Ellis (bass), and Joel Amey (drums). .....
\\
Action 3: Lookup[Wolf Alice instruments]
Observation 3: No more results.

\textbf{Finish[Wolf Alice]} 
Thought 4: I know they reportedly consist of four members and typically feature guitar, bass, drums, and \textcolor{red}{vocals whwereas Badly Drawn Boy only have one member} . (label: Badly Drawn Boy)
\end{tcolorbox}
\caption{Examples of \textit{Reasoning Error}. The incorrect part is highlighted in \textcolor{red}{red}.}
\label{prompt:A}
\end{figure*}
\begin{figure*}
\begin{tcolorbox}[colback=blue!5!white,colframe=blue!75!black,title= ,width=\textwidth]
Answer the following question: \{\{questions\}\} ,\\ with following documents: \{\{documents\}\}. \\
Your response should strictly follow the format:\\
Explanations :[give your step by step Analysis here ]\\
\\
FINAL:(BE CONCISE, ONLY a FEW phrases)\\

let's think step by step
\end{tcolorbox}
\caption{Final answer generation prompt for multi hop QA}
\end{figure*}
\begin{figure*}
\begin{tcolorbox}[colback=blue!5!white,colframe=blue!75!black,title= ,width=\textwidth]
You're a disambiguation expert analyzing "\{\{phrase\}\}" in: 
\{\{self.question\}\}  
Instruction:\\
1. Analyze the question by considering these potential ambiguities:\\
   - Temporal: Check for unclear time references, periods, or temporal scope\\
   - Entity: Identify names, references, or terms that could refer to multiple entities\\
   - Semantic: Look for words with multiple meanings (polysemy/homonymy)\\
   - Scope: Consider possible boundaries and levels of detail\\
   - Intent: Examine possible purposes and expected answer types\\
   - Cultural: Consider cultural-dependent interpretations\\
   - Quantitative: Check for unclear measurements or numerical references\\
   - Linguistic: Analyze syntax and referential clarity\\
   - Categorical: Consider possible classification schemes\\
   - Contextual: Examine required background knowledge and relationships\\
2. Analyze the question word by word. Return disambiguated question and its interperatation for each different meaning\\ 
\\
Here is what we currently know
Documents:\{\{context\}\}\\
\\
pick top 5 questions that are best in disambiguating the question. (covers different meanings of the questions) and strictly FOLLOW the format: response: question1; question2;....
\end{tcolorbox}
\caption{Subcomponent Question generation prompt for Ambiguous QA}\label{prompt:B}
\end{figure*}

\begin{figure*}
\begin{tcolorbox}[colback=blue!5!white,colframe=blue!75!black,title= ,width=\textwidth]
The question may be ambiguous and have multiple correct answers, and in that case, you have to provide a long-form answer including all correct answers.\\
1. Carefully go through all the given documents.\\
2.The using your and context, provide answer. \\
Your response should strictly follow the format:\\
Explanations (Step 2):[give your step by step Analysis here ]\\
FINAL(Step 2):\\
Please ONLY reply according to this format
\\
Question: \{\{questions\}\} 
Document: \{\{documents\}\}
let's think step by step
\end{tcolorbox}
\caption{Final answer generation prompt for Ambiguous QA}\label{prompt:C}
\end{figure*}
\begin{figure*}
\begin{tcolorbox}[colback=blue!5!white,colframe=blue!75!black,title= ,width=\textwidth]
Answer the \{\{questions\}\} based on on the document info. For each question find as many answers as possible. Response all the answers in a short paragraph (as specific as possible). \\
Relevant Document: \{\{context\}\}
\end{tcolorbox}
\caption{Subcomponent question answering prompt}\label{prompt:D}
\end{figure*}

\begin{figure*}
\begin{tcolorbox}[colback=blue!5!white,colframe=blue!75!black,title=example Output of TreeRare ,width=\textwidth]
\textbf{Question:} What screenwriter with credits for `Evolution' co-wrote a film starring Nicolas Cage and Téa Leoni?\\
\textbf{Dependency Parser Output:}\begin{verbatim}
ROOT []
|   co-wrote --root--> [What screenwriter with credits for
"Evolution" co-wrote a film starring
Nicolas Cage and Tea Leoni]
|  |-- screenwriter --nsubj--> [screenwriter with credits for "Evolution"]
|  |  |  |-- What --det--> [What]
|  |  |  -- credits --nmod--> [credits for "Evolution"]
|  |  |     |  |-- with --case--> [with]
|  |  |     |  -- Evolution --nmod--> [Evolution]
|  |-- film --obj--> [film starring Nicolas Cage and Tea Leoni]
|  |  |  |-- a --det--> [a]
|  |  |  -- starring --acl--> [starring Nicolas Cage and Tea Leoni]
|  |  |     |  -- Nicolas Cage --obj--> [Nicolas Cage and Tea Leoni]
|  |  |     |     |  -- Tea Leoni --conj--> [Tea Leoni]
|  -- ? --punct--> [?]
\end{verbatim}
\end{tcolorbox}
\caption{Output of \ours{} at question decomposition stage}
\label{output1}
\end{figure*}
\begin{figure*}
\begin{tcolorbox}[colback=blue!5!white,colframe=blue!75!black,title=example Output of TreeRare ,width=\textwidth]
\textbf{Node: credits for `Evolution'\,''} \textbf{Generated Questions:}  Who wrote the screenplay for the film Evolution''? What movie starring Nicolas Cage and Tea Leoni was co-written by the same screenwriter as Evolution''? What year was Evolution'' released?
\textbf{Subcomponent Question Answering:}
The screenplay for the film Evolution'' was co-written by David Diamond and David Weissman, along with a story by Don Jakoby. The same screenwriters, Diamond and Weissman, also co-wrote the film The Family Man,'' which stars Nicolas Cage and Tea Leoni. ``Evolution'' was released in the year 2001.
\\\textbf{Node: screenwriter with credits for `Evolution'\,''} \textbf{Generated Questions:} Who are the screenwriters credited for Evolution''? What is the title of the film starring Nicolas Cage and Tea Leoni co-written by the same screenwriters? In what year was ``Evolution'' released?
\textbf{Subcomponent Question Answering:}
The screenwriters credited for Evolution'' are David Diamond and David Weissman. They also co-wrote the film The Family Man,'' which stars Nicolas Cage and Téa Leoni. ``Evolution'' was released in the year 2001.
\\\textbf{Node: Nicolas Cage and Téa Leoni''} \textbf{Generated Questions:}  What movies have Nicolas Cage and Téa Leoni starred in together? Who co-wrote the screenplay for the film Evolution''? What are the main themes of the film ``Evolution''?
\textbf{Subcomponent Question Answering:}
Nicolas Cage and Téa Leoni starred together in the film The Family Man,'' a romantic comedy-drama released in 2000. The screenplay for Evolution'' was co-written by David Diamond and David Weissman, in collaboration with Don Jakoby, who originally came up with the story. The main themes of ``Evolution'' include the unintended consequences of technological advancement and the humorous exploration of evolution itself, particularly focusing on alien organisms and how they adapt to Earth, leading to unexpected and comedic situations.
\\\textbf{Node: starring Nicolas Cage and Téa Leoni''} \textbf{Generated Questions:} What film did Nicolas Cage and Téa Leoni star in together? Who were the co-writers of the screenplay for Evolution''? What are the main themes explored in the film Evolution''?  When was The Family Man'' released? What genre does the film ``The Family Man'' belong to?
\textbf{Subcomponent Question Answering:}
Nicolas Cage and Téa Leoni starred together in the film The Family Man,'' which was released in 2000. The screenplay for Evolution'' was co-written by David Diamond and David Weissman. The film Evolution'' explores themes such as the rapid development and adaptation of species, the impact of science and technology on society, and environmental issues tied to evolution. The Family Man'' was released on December 22, 2000, and it belongs to the romantic comedy-drama genre.
\\\textbf{Node: film starring Nicolas Cage and Téa Leoni''} \textbf{Generated Questions:}What film did Nicolas Cage and Téa Leoni star in together? Who were the co-writers of the screenplay for Evolution''?When was ``The Family Man'' released?
\textbf{Subcomponent Question Answering:}
Nicolas Cage and Téa Leoni starred together in the film The Family Man,'' which was released in 2000. The screenplay for Evolution'' was co-written by David Diamond and David Weissman, who frequently collaborate on various projects. ``The Family Man'' premiered on December 22, 2000.
\\\textbf{Final Answer:}
The screenwriters credited for the film Evolution'' are David Diamond and David Weissman.
\end{tcolorbox}
\caption{Output of \ours{} at the Subcomponent Question Answering Stage}
\label{output2}
\end{figure*}

\end{document}